\documentclass[letterpaper, 10 pt, conference]{ieeeconf}  % Comment this line out if you need a4paper

\IEEEoverridecommandlockouts                              % This command is only needed if 
\usepackage{graphicx} % for pdf, bitmapped graphics files
\usepackage{amsmath} % assumes amsmath package installed
\usepackage{amssymb}  % assumes amsmath package installed
\usepackage{tabularx}
\usepackage{booktabs}
\usepackage{caption}
\usepackage{subcaption}
\usepackage{hyperref}

\title{\LARGE \bf
Whenever, Wherever: Towards Orchestrating Crowd Simulations with Spatio-Temporal Spawn Dynamics
}

\author{Thomas Kreutz$^{1}$, Max Mühlhäuser$^{1}$ and Alejandro Sanchez Guinea$^{2}$%
\thanks{$^{1}$Telecooperation Lab at the Technical University Darmstadt, Germany, {\tt\small \{kreutz@tk, max@informatik\}.tu-darmstadt.de}, \indent $^{2}$NTT DATA, Luxembourg, {\tt\small alejandro.guinea@global.ntt}}
}

\begin{document}

\maketitle
\thispagestyle{empty}
\pagestyle{empty}

%%%%%%%%%%%%%%%%%%%%%%%%%%%%%%%%%%%%%%%%%%%%%%%%%%%%%%%%%%%%%%%%%%%%%%%%%%%%%%%%

\begin{abstract}

%Realistic crowd simulations are crucial for immersive virtual environments, requiring both individual behaviors (microscopic dynamics) and overall crowd patterns (macroscopic characteristics). While recent data-driven approaches like deep reinforcement learning excel at microscopic realism, they often overlook macroscopic features, such as crowd density and flow. A key macroscopic element is the spatio-temporal arrival dynamics of agents, including their spawn and destination locations. Traditional methods, like random placements or fixed schedules, lack diversity and realism. To address this, we introduce a new crowd simulation concept that integrates macroscopic elements with microscopic techniques, specifically using Neural Temporal Point Processes (nTPP) to model temporal arrival patterns and a Gaussian Mixture Model (GMM) for destination locations. This nTPP-GMM approach enhances realism by learning from real-world data and sampling from a spatio-temporal distribution of agent arrivals and destinations. We incorporate this into a crowd simulation framework with imitation learning for pedestrian dynamics and evaluate it using three diverse real-world datasets. Our method not only simulates realistic scenes but also allows for scenario customization, making it a valuable tool for crowd analysis and scenario generation.

Realistic crowd simulations are essential for immersive virtual environments, relying on both individual behaviors (microscopic dynamics) and overall crowd patterns (macroscopic characteristics). While recent data-driven methods like deep reinforcement learning improve microscopic realism, they often overlook critical macroscopic features such as crowd density and flow, which are governed by spatio-temporal spawn dynamics, namely, when and where agents enter a scene. Traditional methods, like random spawn rates, stochastic processes, or fixed schedules, are not guaranteed to capture the underlying complexity or lack diversity and realism. To address this issue, we propose a novel approach called nTPP-GMM that models spatio-temporal spawn dynamics using Neural Temporal Point Processes (nTPPs) that are coupled with a spawn-conditional Gaussian Mixture Model (GMM) for agent spawn and goal positions. We evaluate our approach by orchestrating crowd simulations of three diverse real-world datasets with nTPP-GMM. Our experiments demonstrate the orchestration with nTPP-GMM leads to realistic simulations that reflect real-world crowd scenarios and allow crowd analysis.

\end{abstract}

%%%%%%%%%%%%%%%%%%%%%%%%%%%%%%%%%%%%%%%%%%%%%%%%%%%%%%%%%%%%%%%%%%%%%%%%%%%%%%%%

\section{Introduction}

Crowd simulations are a key ingredient for building immersive virtual environments with artificial agents or realistic training environments for mobile robots to be deployed in public places~\cite{ling2024socialgail}. While mobile robots can benefit society by being used to deliver goods, guide visitors, or serve as mobile trashcans~\cite{robla2017working, mukherjee2022survey}, one critical challenge is their safe navigation through a crowded human environment~\cite{kruse2013human}. In this context, realistic crowds are crucial so that mobile robots can leverage them as training environments to learn appropriate and safe navigation policies~\cite{ling2024socialgail}.

To simulate realistic crowds, recent advances in data-driven crowd simulation methods that are based on deep reinforcement learning (e.g., \cite{lee2018crowd, panayiotou2022ccp}), imitation learning (e.g., \cite{qiao2019scenario, ling2024socialgail}), or their combination (e.g., \cite{charalambous2023greil}), greatly outperform classical methods (e.g., the Social Force Model~\cite{helbing1995social}, or ORCA~\cite{van2011reciprocal}) in terms of microscopic crowd realism by learning to imitate a large variety of realistic behaviors, such as goal-seeking, collision avoidance, or grouping.

While existing methods for data-driven crowd simulation focus strongly on enhancing the microscopic realism of individual agent behaviors, learning macroscopic features, such as agent spawn dynamics, remain underexplored. Spatio-temporal spawn dynamic of agents, including (i) \textit{where} agents enter and exit a scene, and (ii) \textit{when} agents appear in a scene, directly influence crowd density and flow~\cite{he2020informative}. In this work, we focus on the temporal spawning of agents, which is traditionally handled by fixed or random spawn rates~\cite{dickinson2019virtual, de2023exploring, karamouzas2009density, kraayenbrink2014semantic, panayiotou2022ccp}, learned with traditional stochastic process~\cite{zhou2012understanding, zhong2015learning, he2020informative}, or follows pre-defined schedules~\cite{liu2018social, bein2020simulating, charalambous2023greil}. Random or fixed spawn rates as well as fixed-schedules lack diversity and realism, while traditional stochastic processes, such as the Poisson process, are limited in their ability to capture complex temporal interdependencies~\cite{shchur2021neural}. Furthermore, while the temporal spawn dynamics of agents is an important macroscopic characteristic of realistic crowds, %it has not received much attention by the field and is not even considered an essential aspect for crowd authoring~\cite{lemonari2022authoring}. 
it has not received the necessary attention from previous works to the point of overlooking it when considering crowd simulation components~\cite{lemonari2022authoring}.

In this paper, we propose an approach to learn orchestrating the initialization of agents in crowd simulations in space and time. More specifically, our goal is to learn spatio-temporal spawn dynamics from real-world data, i.e., \textit{when} and \textit{where} agents appear in a scene and \textit{where} they will go. Our approach learns spawn timings based on neural Temporal Point Processes (nTPPs) that are combined with a spawn-conditional Gaussian Mixture Model (GMM) for spawns and goals of each agent to an nTPP-GMM. The nTPP-GMM jointly models a spatio-temporal spawn and goal distribution, which can be used to orchestrate a crowd simulation (i.e., to dynamically control the initialization and flow of agents by sampling from the spatio-temporal distribution). %This orchestration ensures that the simulated crowd behaviors and patterns are realistic, reflecting the natural flow and density variations observed in real-world crowd scenarios.

%, which specifies  \textit{when} and \textit{where} agents appear in the simulation, as well as \textit{where} they are headed

% To learn where common spawns and goals of agents are, we cluster the respective start and end positions of each trajectory, and then compute a co-occurrence distribution between the resulting clusters, which leads to a spawn-conditional GMM. To learn when and where agents spawn in a scene and where they will go, we propose a neural Temporal Point Process Gaussian Mixture Model (nTPP-GMM), which effectively models the spatio-temporal spawn dynamics of a crowd

%called nTPP-GMM that models
%spatio-temporal spawn dynamics using Neural Temporal Point
%Processes (nTPPs) that are coupled with a spawn-conditional
%Gaussian Mixture Model (GMM) for agent spawn and goal %positions.

%Learning the temporal spawn of agents with nTPPs surpasses conventional models like Poisson processes~\cite{zhou2012understanding, zhong2015learning}, which leads to more realistic macroscopic aspects of crowd simulation.

Our approach is evaluated by orchestrating a crowd simulation framework that uses imitation learning to learn and model individual agent behavior. We evaluate our method on three real-world datasets, including Grand Central Station~\cite{zhou2012understanding, yi2015understanding}, Edinburgh Forum~\cite{majecka2009statistical}, and ETH University~\cite{pellegrini2010improving}. Our results demonstrate that the orchestration of a crowd simulation with nTPP-GMM can effectively replicate realistic crowd scenarios and allows us to implicitly analyze crowd behaviors, such as specific situations or crowd flows, solely by simulation. Our contributions are:
\begin{itemize}
\item nTPP-GMM, a novel approach for learning spatio-temporal spawn dynamics from real-world data. 
\item Orchestration of simulated crowds: Using nTPP-GMM, we can orchestrate crowd simulations, which leads to realistic crowd flows and crowd scenarios.
\end{itemize}

\section{Related Work}

\subsection{Temporal Pedestrian Initialization}
\textit{When} to initialize new agents in a crowd simulation is a crucial macroscopic feature, which directly influences crowd density and implicitly leads to common crowd flows~\cite{he2020informative}. However, learning the spawn dynamics of agents has not received much attention by prior works and is not considered an important aspect for authoring crowds~\cite{lemonari2022authoring}. %In a crowd simulation setup, after agents are spawned, they usually have a single goal or work through an agenda~\cite{lemonari2022authoring}. 
Common agent spawning techniques include a fixed spawn rate~\cite{dickinson2019virtual, de2023exploring, karamouzas2009density} or a random distribution coupled with conditions~\cite{kraayenbrink2014semantic}, spawn schedules taken from real data~\cite{liu2018social, bein2020simulating, nishida2023crowd}, or stochastic processes~\cite{zhou2012understanding, zhong2015learning, he2020informative}. While classical stochastic processes, such as the Poisson process as used in~\cite{zhou2012understanding, zhou2015learning} or HDP in~\cite{he2020informative}, can capture relatively simple patterns, these approaches can not learn complex dependencies between event occurrences~\cite{shchur2021neural}. To address this gap for crowd simulations, we propose learning the stochastic spawn process of agents with neural Temporal Point Processes (nTPPs). Unlike classical (non-neural) TPPs like the Poisson process, nTPPs can learn complex (spatio-)temporal dependencies to generate a large variety of patterns such as global trends, burstiness, or repeating sub-sequences~\cite{shchur2021neural}, which are essential characteristics needed to realistically simulate the spawn dynamics of agents in crowd simulations.

\subsection{Understanding Crowds with Data-Driven Simulation}
Understanding crowds involves capturing both macroscopic aspects, such as agent spawns and destinations, and microscopic behaviors, like common agent paths. Previous works, such as those in~\cite{zhou2012understanding, zhong2015learning, he2020informative}, use classical statistical methods, like mixture models, Poisson processes, and hierarchical Dirichlet processes (HDP), to model pedestrian flows and patterns. These methods require additional methods for simulation, such as extracting path patterns, incorporating extra collision avoidance measures~\cite{zhong2015learning}, or relying on separate crowd simulation frameworks that require specific parameters as an input to control each agent~\cite{he2020informative}. Although effective, prior works are constrained by their specific assumptions about the data distribution. In contrast, our paper leverages deep learning-based methods for crowd simulation %and learning of spatio-temporal spawn dynamics from real-world data without such limitations, 
which eliminates any assumptions about the data distribution, while allowing implicitly analysis of the underlying crowd data through simulation. %This eliminates the need for separate rule-based parametric simulation frameworks or additional methods for pattern extraction and collision avoidance. %Instead, through crowd orchestration, we can implicitly analyze the underlying crowd data through simulation. %Deep learning's ability to learn complex, data-driven patterns allows us to capture more realistic crowd dynamics and interactions, facilitating implicit crowd analysis through integrated simulation.

%In contrast, our approach employs deep learning to learn spawn dynamics and agent behaviors directly from real-world data. This eliminates the need for separate simulation frameworks or additional methods for pattern extraction and collision avoidance. Deep learning's ability to learn complex, data-driven patterns allows us to capture more realistic crowd dynamics and interactions, facilitating implicit crowd analysis through integrated simulation.

%While all these previous methods have been effective, they are limited in capturing the underlying crowd behavior due to their specific assumptions about the data distribution. In comparison, our approach uses deep learning to learn spawn dynamics and agent behavior directly from real-world data and does not make any assumptions about the underlying data distribution. This allows for learning more realistic spawn dynamics, agent behavior, and interactions between agents and their environment, which allows us to analyze the underlying crowd implicitly by simulation. %By automatically learning from data, our method can lead to more realistic and flexible simulations compared to conventional approaches.

\subsection{Learning Agent Policies for Crowd Simulation}

Crowd simulation has progressed from more traditional rule-based approaches such as Reynolds' flocking algorithm~\cite{reynolds1987flocks}, Helbing's Social Force Model~\cite{helbing1995social}, or Van den Berg's Optimal Reciprocal Collision Avoidance (ORCA)~\cite{van2011reciprocal} to data-driven approaches that improve the realism of individual agent behavior in a crowd. Recent data-driven methods focus strongly on advancing the realism and controllability of crowd simulations. These methods span imitation learning using on-policy (deep RL, IRL)~\cite{lee2018crowd, panayiotou2022ccp, qiao2019scenario, charalambous2023greil, ling2024socialgail}, off-policy methods (e.g., BC)~\cite{qiao2019scenario}, but also generative methods for multi-agent simulation that are mainly driven by trajectory prediction over a limited time horizon~\cite{sadeghian2019sophie, kosaraju2019social, mangalam2021goals, yue2022human, rempe2022generating, shi2022social, rempe2023trace, chiara2022goal}.

\section{Approach}

\subsection{Problem Statement and Overview}

\begin{figure*}[t]
    \centering
    \includegraphics[width=.9\linewidth]{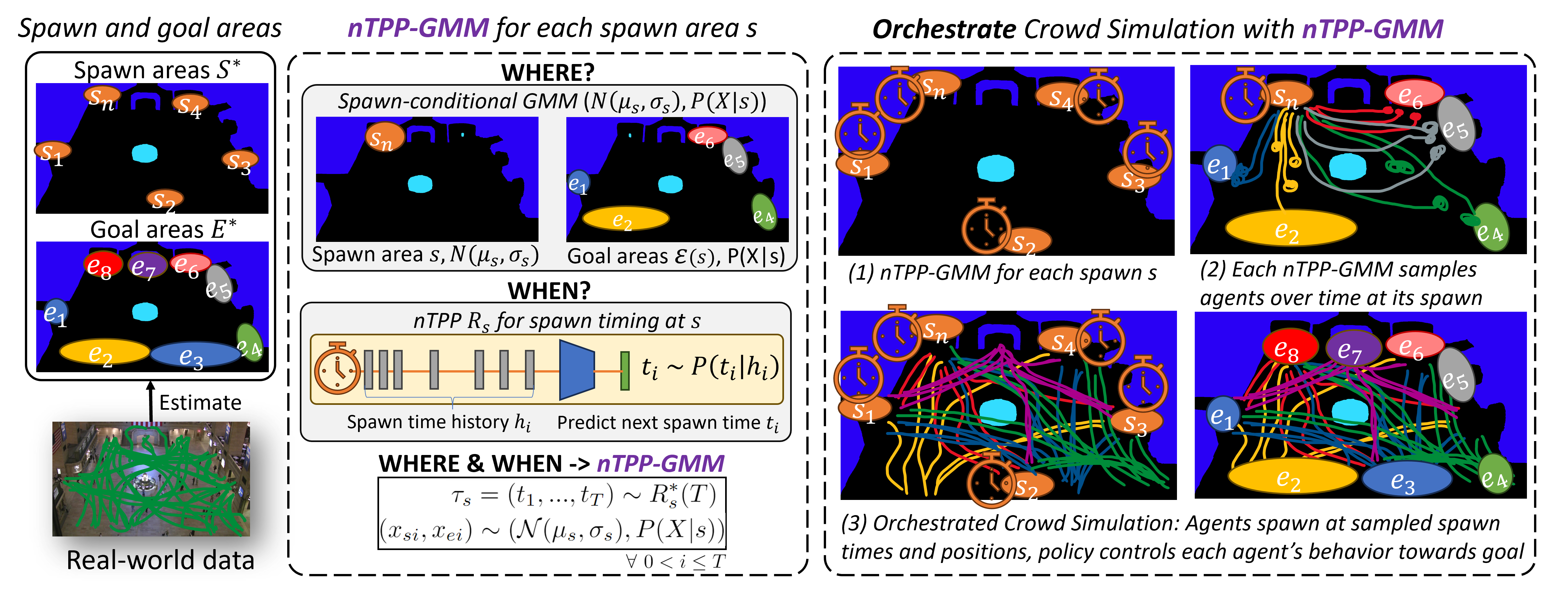}
    \caption{\small We propose orchestrating a crowd simulation with spatio-temporal spawn dynamics learned by an nTPP-GMM from real-world data. First, spawn and goal areas $S^{*}, E^{*}$ are estimated from real-world data defining \textit{where} they spawn and \textit{where} their goal is. Second, for each spawn $s\in S^{*}$, we train an nTPP-GMM to learn \textit{when} agents will spawn, which allows sampling a spawn time sequence $\tau_{s}$ for simulation. Each spawn time $t_{i} \in \tau_{s}$ yields a new agent with sampled spawn and goal positions $(x_{si}, x_{ei})$. Third, to orchestrate a crowd simulation, the nTPP-GMM of each spawn independently generates agents (1,2), and a policy controls each agent while they are present in the scene to reach their goal (3).}
    \label{fig:approach_figure}
\end{figure*}

Our work addresses the challenge of learning spatio-temporal spawn dynamics from real-world crowd trajectory data. Specifically, we aim to determine \textit{when} and \textit{where} agents spawn in a scene, as well as \textit{where} their goal positions are. Figure~\ref{fig:approach_figure} outlines our approach. First, we cluster the start and end positions of each trajectory to estimate common spawn and goal areas (\(S^{*}\) and \(E^{*}\)). We also compute a co-occurrence distribution between these clusters, leading to a spawn-conditional Gaussian Mixture Model (GMM), which jointly defines \textit{where} agents spawn and \textit{where} their goal is. To model \textit{when} agents spawn, we combine this GMM with a neural Temporal Point Process (nTPP), forming an nTPP-GMM that learns the timing of spawns for each spawn area. This allows us to generate a sequence of spawn times, where each time generates a new agent with corresponding spawn and goal positions. Finally, a set of nTPP-GMMs orchestrates a crowd simulation by generating agents over time while the agent policy of the simulation moves them toward their goals. %Our approach effectively captures both the spatial distribution of spawn and goal areas and the temporal dynamics of agent spawns, enabling more realistic crowd simulations.

\subsection{Where: Spawn and Goal Areas of Agents}
Agents that traverse public places usually have common spawn and goal areas~\cite{zhou2012understanding}, which are fundamental for a realistic crowd simulation. We are given a dataset of trajectories $D := \{ tr_{1}, ... tr_{N} \}$ obtained from, e.g., object tracks in a real-world scene, where each trajectory $tr_{i} \in D$ is a sequence of $(x,y) \in \mathbb{R}^{2}$ positions of arbitrary length. Our goal is to obtain two sets of clusters for common spawn and goal areas of $D$. To this end, we first split $D$ into a set $S := \{ s_{1}, ... , s_{N} \} \subseteq \mathbb{R}^{2}$ and $E := \{ e_{1}, ..., e_{N} \} \subseteq \mathbb{R}^{2}$, where each $s_{i} \in S, e_{i} \in E$ are the respective start and end positions of a trajectory $tr_{i}$. Next, we apply a clustering algorithm (DBSCAN~\cite{ester1996density}) on $S$ and $E$ separately, to obtain a set of spawn areas $S^{*}$ and goal areas $E^{*}$. For each area in $S^{*}, E^{*}$, we assume them to be normally distributed and compute the respective parameters $\mu_{k}, \sigma_{k}$ so that we can sample positions from each area. %Our approach does not require a perfect clustering. However, better clustering helps. 

To mimic the distribution of agents that move from a spawn area $s \in S^{*}$ to some goal area $e \in E^{*}$, we compute the frequency of co-occurences for all $(s,e) \in S^{*} \times E^{*}$ with a function \mbox{$freq : S^{*} \times E^{*} \rightarrow \mathbb{N}$}. Afterward, given a spawn area $s$, we obtain a probability distribution $P(E | s)$ over the possible goal areas of an agent by computing the relative frequency of each co-occurrence\footnote{This is a simplifying assumption since we assume no information about the real intentions to be available.}. For any $(s,e) \in S^{*} \times E^{*}$, $P(E^{*}=e | s)$ is defined as follows:

\vspace{-.5cm}

\begin{align}
    P(E^{*}=e | s) = \dfrac{freq(s,e)}{\sum_{e' \in E^{*}} freq(s, e')}
\end{align}

Under $P(E^{*}=e|s)$, for each $s$, there is a set of possible goal areas $\mathcal{E}(s) = \{ e \ | \ P(E^{*}=e|s) > 0 \wedge e \in E^{*} \}$, where $s$ and $e$ co-occur at least once (otherwise $P(e|s) = 0$). As a result, a goal area $e \sim P(E^{*} | s)$ can be sampled with weighted uniform sampling for any $s \in S^{*}$. Given a spawn area $s \in S^{*}$, the sampling procedure for an agent's spawn position $x_{s}$ and a goal area $e$ with goal position $x_{e}$ is:
\begin{align}
     x_{s} \sim \mathcal{N}(\mu_{s}, \sigma_{s}), \ 
    x_{e} \sim \mathcal{N}(\mu_{e}, \sigma_{e}) \ \text{with} \ e \sim P(E^{*} | s) 
\end{align}
where $\mu_{s}, \sigma_{s}, \mu_{e}, \sigma_{e}$ are spawn and goal area parameters.

Any goal area $e \in E^{*}$ with $freq(s,e) > 0$ is a Gaussian with $\mu_{e}, \sigma_{e}$ and $e \in \mathcal{E}(s)$. Let $X = \mathbb{R}^{2}$. In this case, the distribution over possible goal positions $P(X = x_{e} | s)$ conditioned on a spawn area $s$ (as described in Equation 2) can more generally be defined as a GMM, which defines the following marginal probability of a goal position $x_{e}$:
\begin{align}
    P(X = x_{e} \ | \ s) = \sum_{k \in \mathcal{E}(s)} \mathcal{N}(X = x_{k} \ | \ \mu_{k}, \sigma_{k}) * \pi^{s}_{k}
\end{align}
where the number of components depends on $s$ and the set of co-occurring goal areas $\mathcal{E}(s)$, $\pi^{s}_{k}$ are the mixture weights, and $\mu_{k}, \sigma_{k}$ are the mixture components, with the interpretation that $\pi^{s}_{k} = P(k | s)$. With Equation 3, we redefine the sampling procedure in Equation 2 as a \textit{spawn-conditional} GMM: 
\begin{align}
    (x_{s},x_{e}) \sim (\mathcal{N}(\mu_{s}, \sigma_{s}), P(X | s))
\end{align}

\subsection{When: Learning spawn timings of agents}

We assume that spawn dynamics of all different spawn areas $s_{i},s_{j} \in S^{*}, i \neq j$ are statistically independent of each other. This assumption allows modeling each spawn area as an independent spawn process. In this context, we propose learning the spawn process at a spawn area $s \in S^{*}$ with a neural Temporal Point Process (nTPP) from real data. Neural TPPs can learn different patterns, such as global trends, burstiness, or repeating subsequences~\cite{shchur2021neural}, which are essential characteristics of agent spawn timings in a crowd. %While a marked nTPP with the interpretation that each spawn is considered a mark would also be a suitable model, the statistical independence assumption of the spawn processes between each spawn motivates the use of a single nTPP for each spawn. 

We build on the implementation\footnote{https://shchur.github.io/blog/2021/tpp2-neural-tpps/} of an nTPP similar to~\cite{du2016recurrent}, which models the nTPP as a recurrent neural network (RNN), which we denote as $R$. Given a sequence of spawn timings \mbox{$\tau = \{t_1, t_2, \dots, t_N\}$}, the model predicts the probability distribution of inter-event times \mbox{$ \Delta t_i = t_{i+1} - t_i $}. To this end, $R$ encodes past inter-event times into a hidden state $ h_i $, which parameterizes the next inter-event time distribution $ f(\Delta t_i \mid h_i) $. $f(\Delta t_i \mid h_i) $ is parameterized as a Weibull distribution, and it represents the likelihood of observing the next inter-event time $ \Delta t_i $ given observed past inter-event times $\{\Delta t_{i-1}, ...\}$. To train $R$ on real data, we split the full sequence $\tau$ into overlapping sliding windows of length $ w $, where each subsequence starts at $ t_k $ and ends at $ t_{k+w} $. Let $ S(T_{end} - t_{k+w}) $ be a survival function that accounts for the probability that no events occur after the last observed event $ t_{k+w} $ in the subsequence, where $ T_{end} $ is the maximum time in the window. The training objective of $R$ is to minimize the negative log-likelihood (NLL) over all sliding windows, which for each subsequence leads to the following loss:
\begin{equation}
    \mathcal{L}_{\text{NLL}} = -\left( \sum_{i=k}^{k+w-1} \log f(\Delta t_i \mid h_i) + \log S(T_{end} - t_{k+w}) \right)
\end{equation}

where $ f(\Delta t_i \mid h_i) $ is the predicted PDF of inter-event times, and $ S(T_{end} - t_{k+w}) $ represents the survival function. 

\subsection{Where+When: nTPP-GMM}
After training the nTPP, we can autoregressively sample agent spawn sequences up to time $T$ from each trained $R_{s}$, which we denote as $R^{*}_{s}(T)$. The sampling is unconditional and starts with a random initial hidden state. By defining each individual spawn $t_{i}$ as a random variable and including the distribution of agent spawn and goal positions from Equation 4, the overall spatio-temporal spawn dynamics of a scene lead to the definition of a \textit{neural Temporal Point Process Gaussian Mixture Model} (nTPP-GMM):

%\begin{align}
%    \tau_{s} \sim R_{s}^{*}(T) \ \text{with} \ \tau_{s} := (t_{1}, ... , t_{T})
%\end{align}

%\begin{align}
%    \tau_{s} = (t_{1}, ..., t_{T}) \sim R^{*}_{s}(T) \\
%    x_{si} \sim \mathcal{N}(\mu_{s}, \sigma_{s}) \\
%    x_{ei} \sim \sum_{k=1}^{K} \pi^{s}_{k} * \mathcal{N}(x_{ei} | \mu_{k}, \sigma_{k}, s) , \ \forall \ 0 < i \leq T
%\end{align}

\vspace{-.6cm}
\begin{align}
    \tau_{s} = (t_{1}, ..., t_{T}) \sim R^{*}_{s}(T) \\
     (x_{si},x_{ei}) \sim (\mathcal{N}(\mu_{s}, \sigma_{s}), P(X | s)), \ \forall \ 0 < i \leq T
\end{align}

where (i) a sequence of spawn timings $\tau_{s} = (t_{1}, ..., t_{T})$ is sampled from the nTPP $R^{*}_{s}(T)$, and (ii) for each spawn time $t_{i} \in \tau_{s}$, a spawn, goal position pair $(x_{si}, x_{ei})$ is sampled from its spawn-conditional GMM defined in Equation 4. %By repeating this process for each spawn $s$, we obtain macroscopic information of all agents in the scene (\textit{where}, \textit{when}). 

%\paragraph{Future Work} The mixture weights and components could also be learned from real data by adding a mixture density network to define a variant of marked ntpps.

\subsection{Orchestrating Crowds with nTPP-GMM}
To orchestrate a crowd simulation, nTPP-GMM is integrated into the initialization process of agents. For each spawn $s \in S^{*}$, the nTPP-GMM samples a set of agents with respective spawn timing, spawn, and goal position according to Equation 6 and 7. With a global clock keeping track of time, agents are spawned at their designated spawn position whenever their spawn time is reached, while the policy of the crowd simulation controls each agent. Finally, agents are removed from the simulation upon reaching the goal position. %To obtain realistic behavior of agents, one we train policy with imitation learning, such as behavior cloning~\cite{bain1995framework, ross2011reduction} or GAIL~\cite{ho2016generative}.

\section{Experiments}

%%%%% FIGURES

\begin{figure}[t]
    \centering
    \smallskip
    \includegraphics[width=\linewidth]{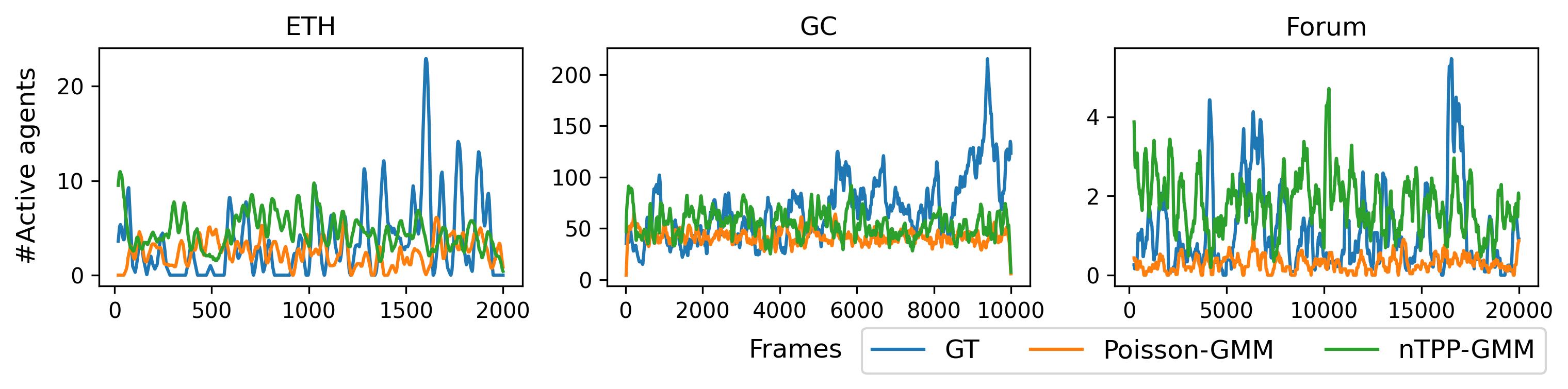}
    \caption{\small Number of agents in the scene (y-axis) at each frame. Our approach approximates the real data better.}
    \label{fig:agents_in_scene_all}
\end{figure}

\begin{figure}[t]
    \centering
        \includegraphics[width=\linewidth]{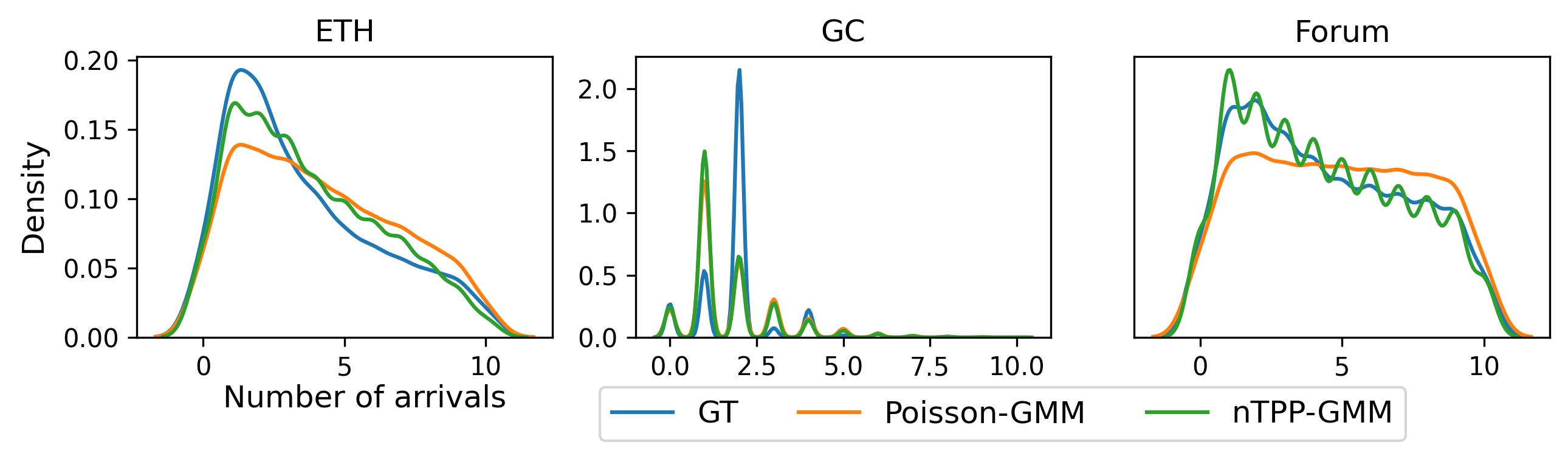}
    \caption{\small Inter-spawn times within 10 frames}
    \label{fig:inter_arrivals_all}
\end{figure}

\begin{figure}[t]
    \centering
    \smallskip
        \includegraphics[width=\linewidth]{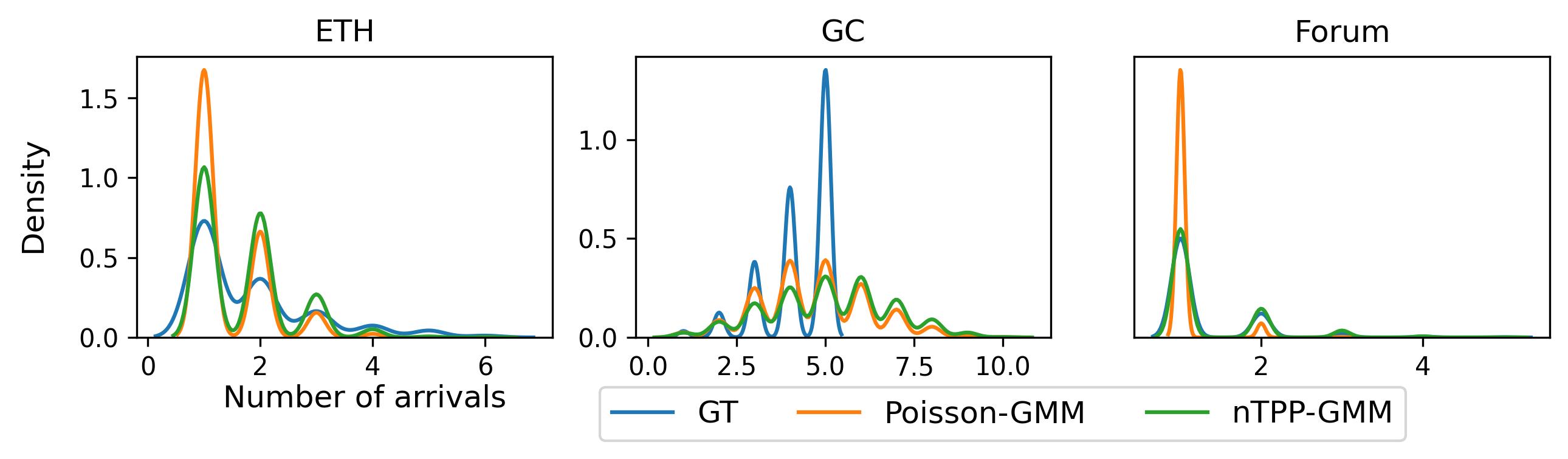}
        \caption{\small Number of spawns within 10 frames}
    \label{fig:num_arrivals_all}
\end{figure}

\begin{figure}[t]
    \centering
    \begin{subfigure}{0.45\linewidth}
        \includegraphics[width=\linewidth]{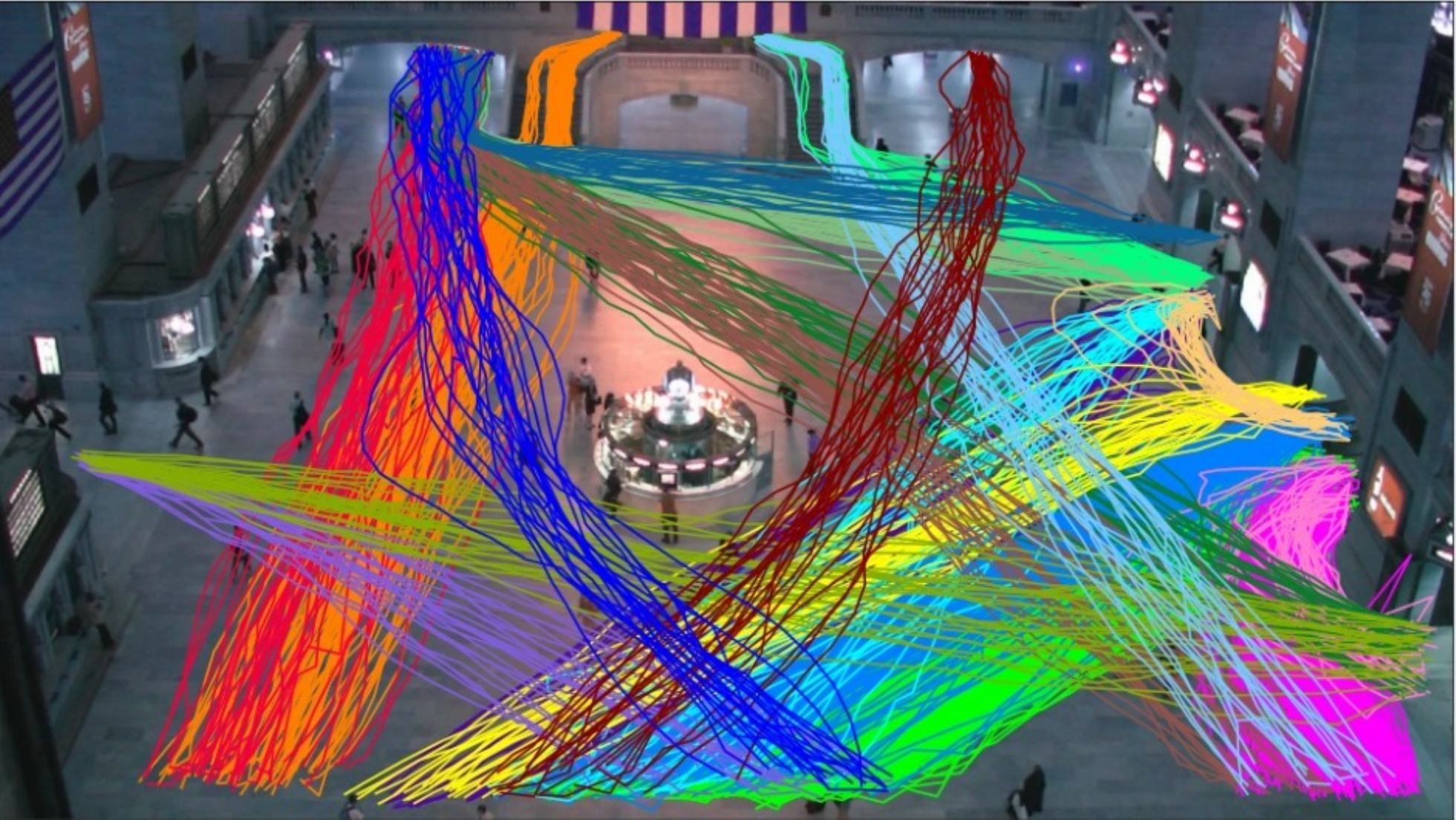}
        \caption{Ground Truth}
        \label{fig:gt}
    \end{subfigure}
    \begin{subfigure}{0.45\linewidth}
        \includegraphics[width=\linewidth]{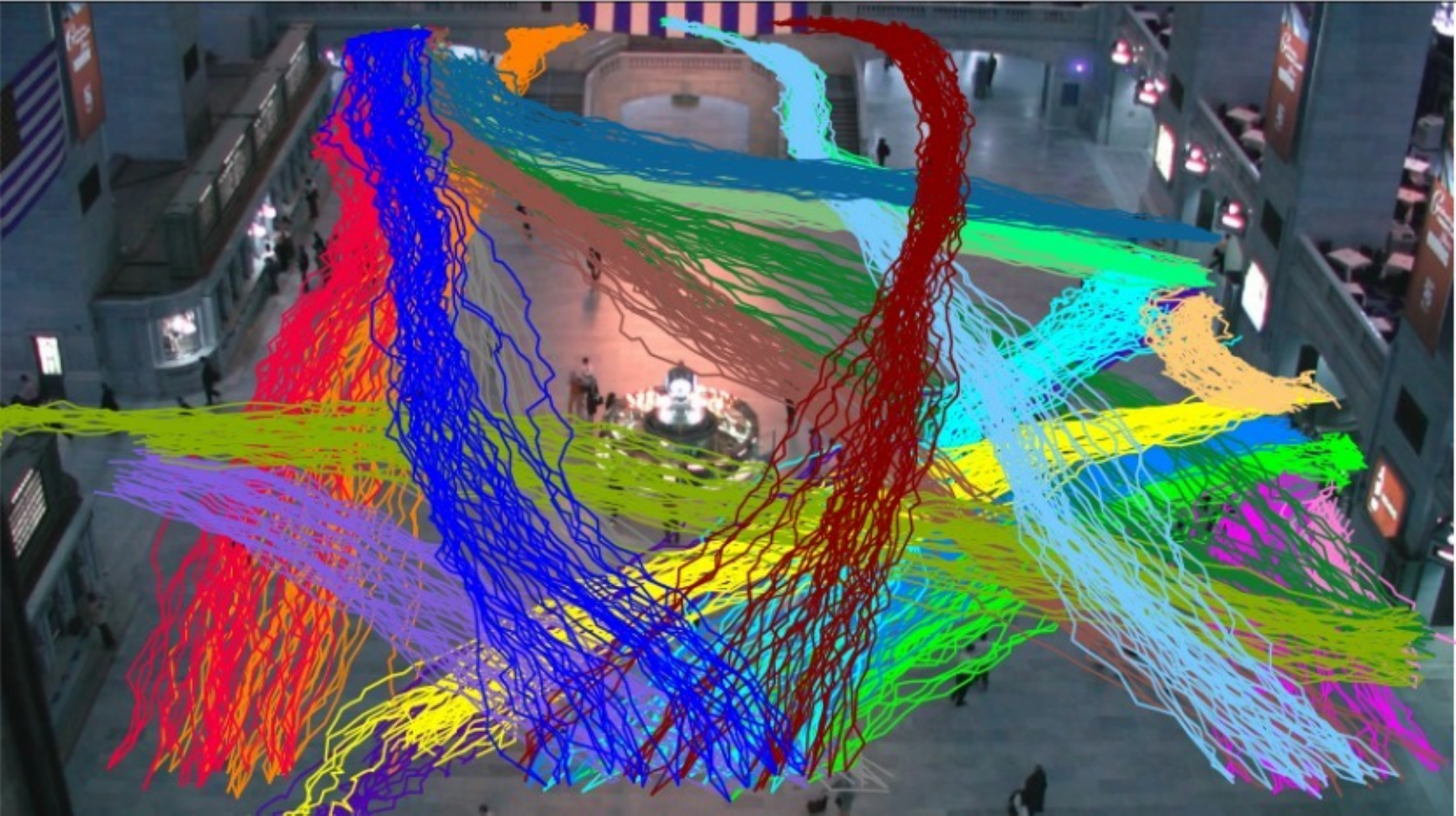}
        \caption{Behavior Cloning}
        \label{fig:pred}
    \end{subfigure}
    \caption{\small Trajectories generated with a behaviour cloning policy (right) effectively approximate the underlying trajectory distribution of the ground truth dataset (left). Each spawn-destination pair is assigned a unique color.}
    \label{fig:bc_vs_realdata}
\end{figure}

\begin{figure}[t]
    \centering
    \smallskip
    \includegraphics[width=\linewidth]{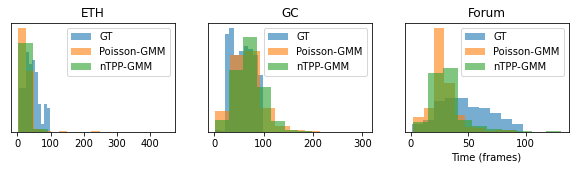}
    \caption{\small Time that agents controlled by our policy spend in the scene in comparison to the real crowd.}
    \label{fig:time_in_scene}
\end{figure}

\iffalse
\begin{figure}[t]
    \centering
    \includegraphics[width=\linewidth]{figures/evaluation/crowd_stats/ntpp_gmm_hyperparams_small.png}
    \caption{We compare hyperparameters for nTPP-GMM against Poisson-GMM in the GC dataset. We train one nTPP-GMM for each hyperparameter combination and sample 5 spawn sequences. We show the mean and two times the standard deviation to visualize the corresponding variety.}
    \label{fig:ntpp_hyperparam_1}
\end{figure}
\fi

\begin{figure}[t]
    \centering
    \includegraphics[width=\linewidth]{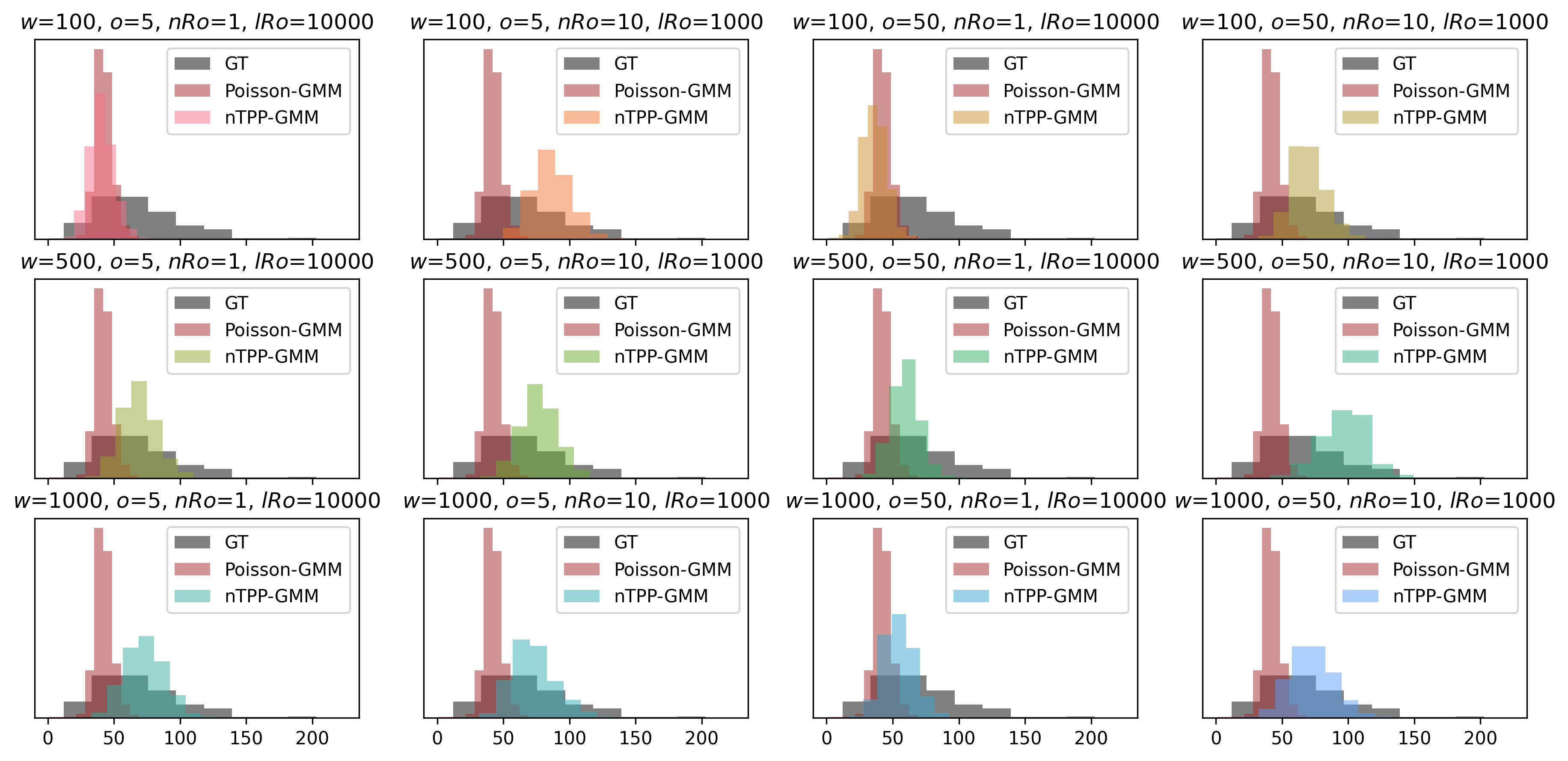}
    \caption{\small In GC, we compare 5 samples of different hyperparameter combinations for nTPP-GMM against 10000 GT frames and Poisson-GMM to test the influence of window size ($w$), sliding window overlap ($o$), number ($nRo$) of rollouts, and length ($lRo$) of rollouts. We show the distribution of the number of agents (x-axis) measured at any point in time.} %We train one nTPP-GMM for each hyperparameter combination and sample 5 sequences. We show the distribution of the number of agents over time as a KDE to emphasize the spread and variety of agents over time, and compare nTPP-GMM to a Poisson process and the ground truth (GT).}
    \label{fig:ntpp_hyperparam_2}
\end{figure}

\begin{table}[t]
    \centering
    \smallskip
    \begin{tabular}{cc}
    
        %\toprule
        Hyperparameter & Values  \\
        \midrule
        Window Size ($w$) & $[100, 500, 1000]$ \\
        Sliding Window Overlap ($o$) & $[5, 50]$ \\ 
        Number Rollouts ($nRo$) & $[1,10]$\\
        Length Rollout ($lRo$) & $[1000, 10000]$ \\
        \midrule
        Total Length & 10000 \\
        
    \end{tabular}
    \caption{\small Summary of hyperparameter evaluation: Window size (w), sliding window overlap (o) during training window creation, number (nRo) and length (lRo) of rollouts.}
    \label{tab:hyperparameters}
\end{table}

\begin{figure}[t]
    \centering
    \includegraphics[width=\linewidth]{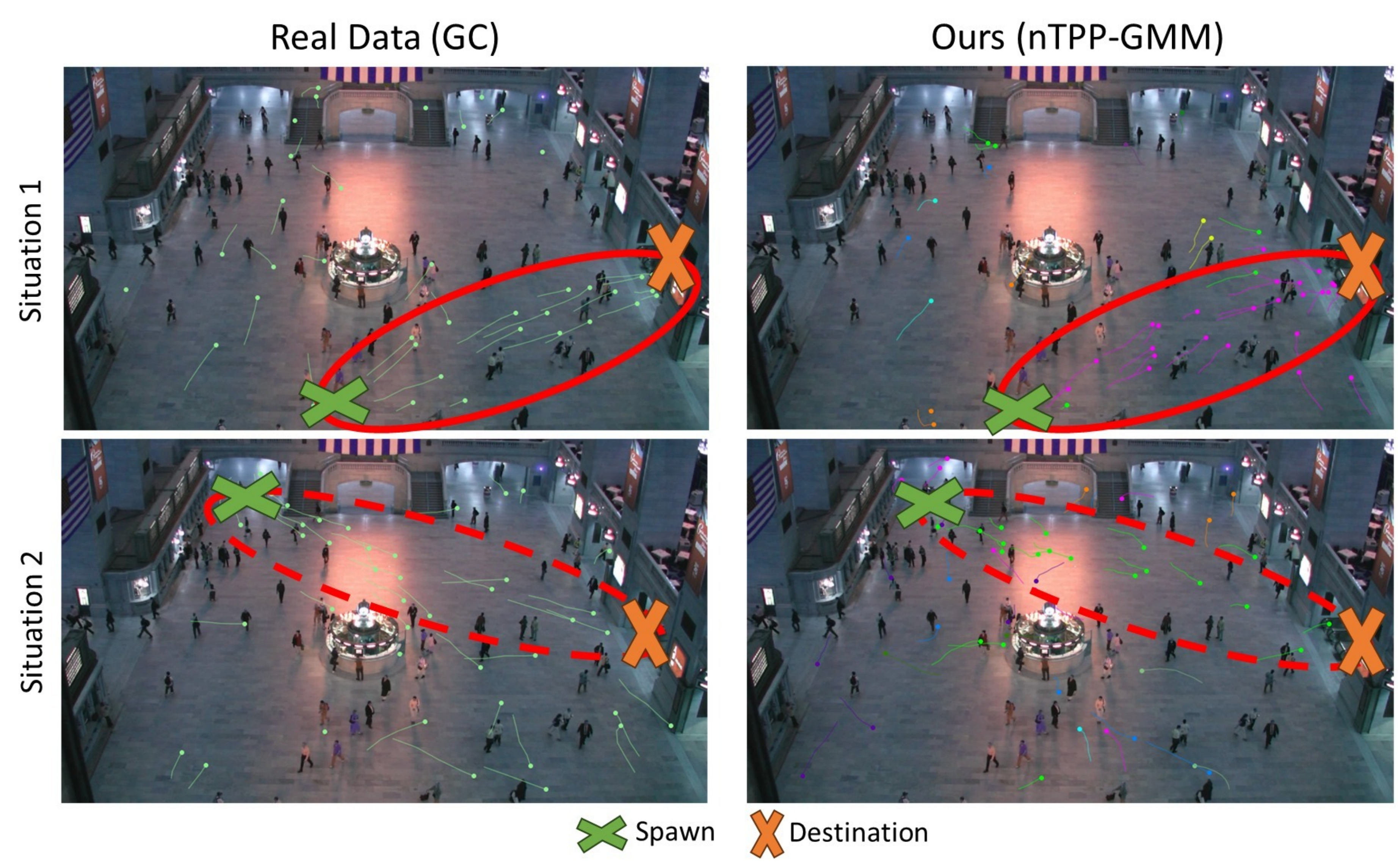}
    \caption{\small Comparison of two crowd flow situations present in the real data in GC that are reproduced by nTPP-GMM.}
    \label{fig:patterns_gc}
\end{figure}

\begin{figure}[t]
    \centering
    \smallskip    
    \includegraphics[width=\linewidth]{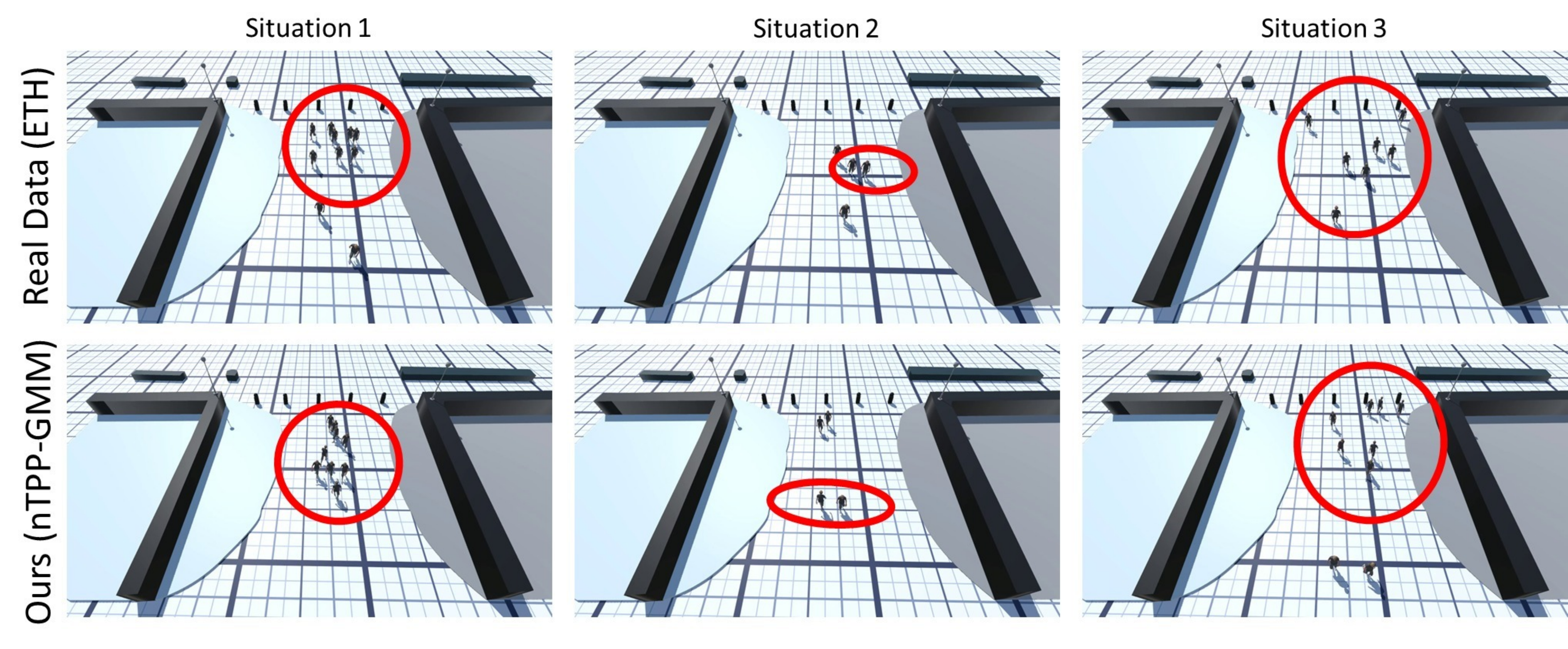}
    \caption{\small Comparison of three scenarios in ETH that differ in crowd density and group sizes reproduced by nTPP-GMM.}
    \label{fig:patterns_eth}
\end{figure}

\begin{figure}[t]
    \centering
    \includegraphics[width=\linewidth]{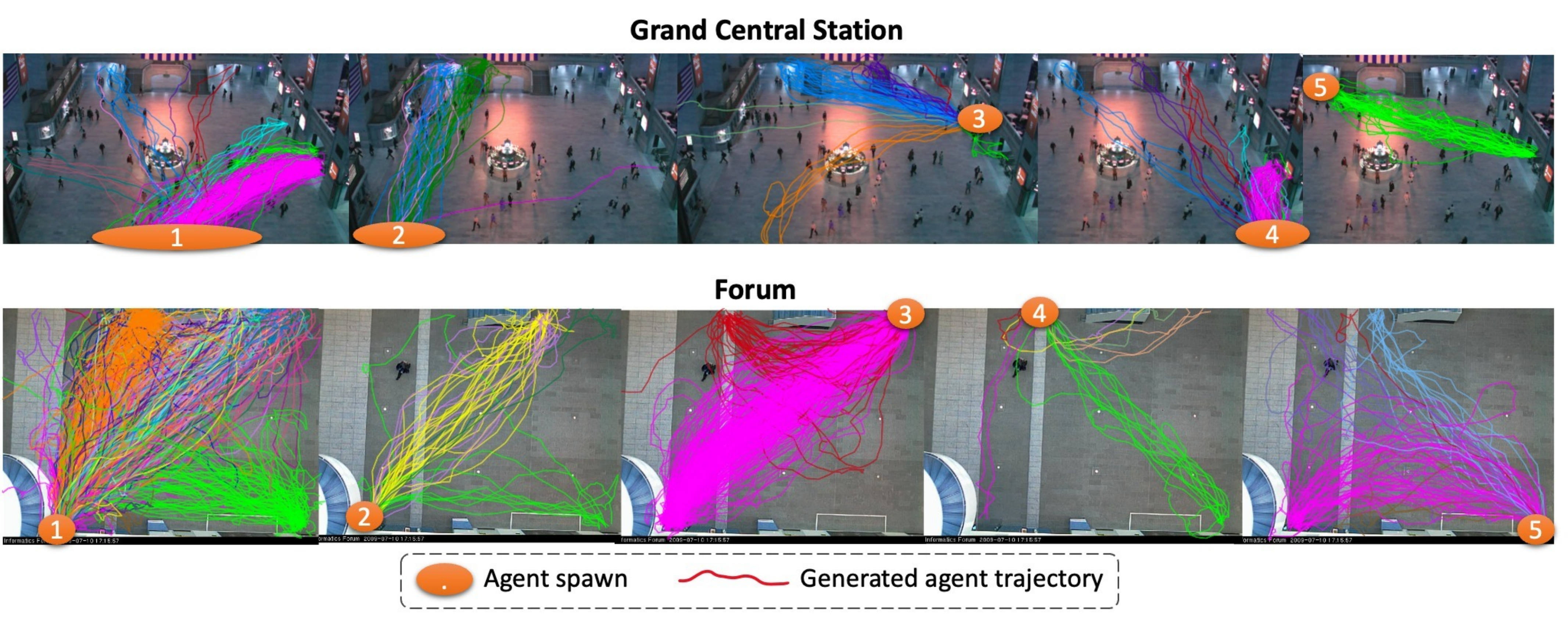}
    \caption{\small Running the simulation in GC and Forum and separating by spawns and destinations implicitly leads to common crowd flows. The crowd density and destination distribution are controlled by the learned nTPP-GMM, while each individual agent is controlled by the policy.}
    \label{fig:analysis_spawns_flows}
\end{figure}

%%%%%% END FIGURES

%\subsection{Overview}
To evaluate our approach, we implemented\footnote{Our code and implementations are available on github: \url{https://github.com/thkreutz/crowdorchestrationsim}} a Gym-based crowd simulation environment~\cite{towers_gymnasium_2023} that can leverage real-world datasets for training of agent policies, and integrate the nTPP-GMM into it. %We evaluate our approach by integrating nTPP-GMM into a crowd simulation framework. 
We compare nTPP-GMM against the ground truth crowd data (GT) and a Poisson process used in ~\cite{zhou2012understanding, zhong2015learning} as a baseline\footnote{To the best of our knowledge, other learning-based methods that learn spawn dynamics from real-world data for crowd simulations not exist in the literature or don't share code (e.g.,\cite{he2020informative}).} (referred to as Poisson-GMM) for learning the spatio-temporal spawn dynamics in three real world scenes. Afterward, we show that our crowd simulation framework combined with nTPP-GMM can realistically simulate real-world scenarios and can be used for crowd analysis. %In our third evaluation, we show that it is possible that we can interactively control the nTPP-GMM to modify the distribution of the simulated crowd by varying, e.g., the destination distribution and the spawn dynamics at each spawn point, which results in different crowd scenarios.

%We have done several crowd video visualizations in the Unity Game Engine as well as overlaying the trajectories by plotting them on top of a reference image at each frame. The supplemented video includes several animated results, which give a better impression on what the final simulation looks like compared to a ground truth.

%Second, we show how our approach can be used to analyze and better understand common crowd flows by using many different real world scene. 
%Finally, we show three types of simulations that are compared to the ground truth: Single-agent simulation in a real crowd, multi-agent simulation with start and goal positions of the dataset, our nTPP-GMM crowd simulation. 

\subsection{Datasets}
We evaluate our crowd orchestration approach in 3 different real-world scenes: Grand Central Station (GC)~\cite{zhou2012understanding, yi2015understanding}, Edinburgh Forum (Forum)~\cite{majecka2009statistical}, and ETH University (ETH)~\cite{pellegrini2010improving}. In Forum, several full days of recordings are available and we chose the recording on July 14, which includes a good amount of agents throughout the whole day as the number of agents in Forum is usually very low. %For Forum, there exist many different days of recordings and usually only a few number of agents are in the scene. We chose the recording on July 14, where there is a good amount of agents throughout the whole day.  
The number of frames and agents in each scene varies significantly between each dataset, which results in differences in terms of agent density, variety of behavior, number of spawns and destinations as well as the spawn dynamics. There are 1929 frames with 258 agents in ETH, 11999 frames with 12255 agents in GC, and 58480 frames with 1331 agents in Forum July 14. 

\iffalse

\begin{table}[t]
    \centering
    \begin{tabular}{ccc}
    \toprule
        Dataset & Frames & Num Agents \\
        \midrule
        GC~\cite{yi2015understanding} & 11999 & 12255 \\
        Forum (Jul14)~\cite{majecka2009statistical} & 58480 & 1331 \\
        ETH University~\cite{pellegrini2010improving} & 1929 & 258
    \end{tabular}
    \caption{Summary over the number of frames and number of agents in each dataset.}
    \label{tab:dataset_stats}
\end{table}
\fi

        %%% gt - poisson - ntpp-gmm
%%%%% ETH 258 - 171 - 316
%%%%% Forum 

\subsection{Implementation Details}

\iffalse
\begin{figure}[t]
    \centering
    \includegraphics[width=\linewidth]{figures/framework.png}
    \caption{Our framework is composed of two main modules. Left (when+where), the spatio-temporal spawn dynamics of agents, destinations, which is decoupled from the agent policy. Right (how), a Gym environment where agents can be trained with imitation learning from arbitrary trajectory datasets in a virtual environment.}
    \label{fig:framework}
\end{figure}
\fi

%A high-level overview of this framework is shown in Figure~\ref{fig:framework}. It is composed of two main modules. The first module implements the required parts for the nTPP-GMM. The second module covers imitation learning of agent policies. 
We use DBSCAN~\cite{ester1996density} to obtain start and goal areas $S^{*}, E^{*}$. The hyperparameters for DBSCAN are optimized for each test scene with $\epsilon = 0.2$ and $min samples=20$ for GC, $\epsilon = 2$ and $min samples=5$ for Forum, and $\epsilon = 0.8$ and $min samples=3$ for ETH. The agent policy is an MLP with two hidden layers of 32 units for all datasets, which matches the policy network used in~\cite{fu2017learning}. We use behavior cloning to train the agent policy for 1000 epochs in our Gym environment with a learning rate of 1e-4 and the Adam optimizer~\cite{kingma2014adam} using the imitation learning library introduced in~\cite{gleave2022imitation}. For the nTPP, we follow the implementation\footnote{https://shchur.github.io/blog/2021/tpp2-neural-tpps/} based on~\cite{du2016recurrent}. The nTPP consists of a GRU with 32 hidden units and an MLP head with 32 units, which is trained for a maximum number of 500 epochs with a learning rate of 1e-4 and the Adam optimizer~\cite{kingma2014adam}. 

%Our gym environment takes an annotated semantic map and the respective trajectories as input to compute the state action pairs for the expert dataset. We obtain a semantic map of Forum and GC by using SAM~\cite{kirillov2023segany} to annotate borders and obstacles. For ETH, we use the semantic map provided by~\cite{mangalam2021goals}. We preprocess each dataset by removing pedestrians that do not move, do not travel very far, and remove all trajectories that are shorter than 20 timesteps or longer than 100 timesteps. %Each dataset is downsampled from its original framerate. ETH is downsampled to 2.5 FPS, GC to 10 FPS, and Forum to 9 FPS. 
%A summary of the number of agents and number of frames is given in Table~\ref{tab:dataset_stats}. We made extensive use of the OpenTraj toolkit~\cite{amirian2020opentraj} to read and preprocess the respective scenes. As a last step, we make sure that the positions in each trajectory are in pixel space by transforming them with their respective homographic transforms.

\subsection{Number of Agents over Time}
We compare the number of agents over time of nTPP-GMM against Poisson-GMM and GT in Figure~\ref{fig:agents_in_scene_all}. Compared to the Poisson-GMM, nTPP-GMM achieves more variety in behavior that is more similar to the ground truth. Our approach can capture different spiking and bursting behavior, which leads to an overall more realistic number of agents and spawn patterns when comparing it to the ground truth. We find that a current limitation of our approach is that strong bursts are not learned in all three datasets (e.g., only in Forum), which we hypothesize is due to the limited amount of training data in GC and ETH compared to Forum. 

\subsection{Inter Spawn Times of Agents}
Figure~\ref{fig:inter_arrivals_all} and Figure~\ref{fig:num_arrivals_all} compare the overall spawn distribution in the scene by comparing inter-spawn times and number of spawns in short 10 frame durations. For the inter-spawn times in Figure~\ref{fig:inter_arrivals_all}, we can see that nTPP-GMM captures an inter-spawn time distribution close to the ground truth that is more accurate than the Poisson-GMM in ETH and Forum. In GC, both nTPP-GMM and Poisson-GMM have a slightly shorter inter-spawn time than the ground truth. For the number of spawns in Figure~\ref{fig:num_arrivals_all}, nTPP-GMM outperforms Poisson-GMM in terms of realism in both ETH and Forum. In GC, there is a difference in burst behavior given that both Poisson-GMM and nTPP-GMM do not replicate the spike at 5 or 4 agents. The spawn behavior in GC is much more complex compared to ETH and Forum, so some spawns might be more difficult to imitate than others. The independence assumption between spawns can also limit the realism in some spawns, which in the real world may have a dependency.

\subsection{Ablation Study on Hyperparameters}\label{sec:ablation}
We evaluate the impact of hyperparameters on nTPP-GMM in GC, which are summarized in Table~\ref{tab:hyperparameters}. We vary between training the nTPP-GMM on sliding windows of short subsequences (100 frames), mid-length subsequences (500 frames), and long subsequences (1000 frames). It is trained on all frames of the GC dataset. We also vary the overlap between the sliding windows during training as well as the length of the final autoregressive rollout. The final rollout is obtained either by doing a rollout with the nTPP-GMM of length 10000 or 10 rollouts of length 1000 that are concatenated to obtain 10000 timesteps. Each rollout strategy is repeated 5 times to get 5 spawn sequence samples for each hyperparameter combination. 

With a short window size of 100, we approximate a Poisson process when sampling rollouts of 10000 frames. Overall, a larger window size of 500 or 1000 is desirable, while a rollout of 1000 or 10000 seems to lead to more similar agent distributions over time.  

\subsection{Verifying Realism of Crowd Simulation Agent Policy}

When integrating nTPP-GMM into a crowd simulation framework, the simulation requires a policy for agents that effectively mimics how long they usually stay inside the scene so that the number of agents remains realistic over time. Otherwise, if agents stay too long, nTPP-GMM will simply continue spawning more agents and the crowd density can (theoretically) grow infinitely. For a proof of concept, we chose behaviour cloning (BC) as an imitation learning policy, which learns these timings as well as a distribution over common paths in the data. The time agents usually spend in the scene is shown in Figure~\ref{fig:time_in_scene}, where we can see that the distribution is close to the real data. That our policy learns to mimic the distribution over trajectories that connect the learned spawn and goal positions is shown in a simple experiment in Figure~\ref{fig:bc_vs_realdata}, where we show several rollouts of the BC policy in GC and compare generated trajectories between common spawn and goal positions (right) with the real trajectories of the dataset (left).

\subsection{Generation of realistic scenarios}
We show that our approach can reproduce real spatio-temporal spawn dynamics that govern different crowd flows without human interference. We simulate crowds for ETH and GC and qualitatively compare it to different scenarios found in the ground truth. %to show that nTPP-GMM can learn to reproduce different patterns of arrival dynamics, which essentially govern the crowd density and its flows. 

We show that nTPP-GMM can reproduce realistic strong crowd flows of GC in Figure~\ref{fig:patterns_gc}. %A common spawn is highlighted at the bottom and top left of the scene (green X), and a common destination is at the right side of the scene (red X). 
Our approach learns two distinct temporal spawn patterns where many agents walk from either the bottom or the top spawn to the destination on the right-hand side. We further show three different situations that nTPP-GMM can learn to reproduce in ETH in Figure~\ref{fig:patterns_eth}. Agents often walk in groups in ETH. %As shown in Situation 1 and Situation 2, these groups can be rather small (e.g., two people) or quite large (e.g., seven people). However, as shown in Situation 3, many people can be in the scene without being in groups. In the bottom row of Figure~\ref{fig:patterns_eth}, we show that our approach learns to reproduce these situations. 
In the first column (Situation 1), we show that our approach replicates a large group of pedestrians. In the second column (Situation 2), a smaller number of agents is generated. Finally, in the third column (Situation 3), many pedestrians who do not belong to a single big group are simulated.
% Given that our approach can reproduce these kinds of crowd densities and group patterns, our model learns to spawn several agents after one another while also varying the spawning process over time to only spawn a small number of agents at longer time intervals.

\subsection{Application: Implicit Crowd Analysis Through Simulation}
In this experiment, we show that combining nTPP-GMM with imitation learning allows the analysis of the common paths that agents usually take to reach their goal \textit{without explicitly clustering the trajectories of the real dataset}. Instead, we implicitly generate the trajectories of each spawned agent. First, in comparison to the real dataset, we can see in Figure~\ref{fig:bc_vs_realdata} that we learn the overall distribution of the trajectories in GC. Second, our approach allows us to analyze a crowd's common crowd flows from each spawn. We visualize the respective analytical results of a simulation in GC and Forum in Figure~\ref{fig:analysis_spawns_flows}, where we can see a subset of common crowd flows of both datasets without clustering the trajectories, but implicitly generating them using nTPP-GMM and simulating the agents with imitation learning. %This is an advantage over existing methods that cluster the trajectories to understand crowd flows, which is not always trivial in large datasets.

\section{Discussion and Future Work}
With nTPP-GMM, we propose a novel approach for learning the spatio-temporal spawn dynamics of crowd scenes. Our experiments demonstrate its potential for realistic simulations and crowd analysis. Furthermore, we want to emphasize that our approach is flexible and agnostic to the specific methods used for each component. For example, agent behavior can be modeled using any kind of imitation learning algorithm or agent policy. Similarly, while we propose using nTPPs to model the spawn dynamics or DBSCAN to identify the spawn and goal areas, our framework is not limited to a particular nTPP or clustering model. %Also, other methods besides DBSCAN can be used to identify the common spawn and goal locations in a scene, which allows to modify the GMM part of the nTPP-GMM.

Future work should focus on integrating nTPP-GMM into existing crowd simulation frameworks or robot training environments for social navigation. Since nTPP-GMM is independent of the agent policy, it can easily enhance the macroscopic realism in existing simulation systems.
Additionally, while our independence assumption leads to the use of separate nTPPs for each spawn, it also leads to the limitation that nTPP-GMM can not explicitly model precise group spawn dynamics. To overcome this limitation, future improvements should explore marked or spatio-temporal nTPPs, which can be trained end-to-end on aspects such as spawns, goals, and the spawn timings in a unified way, which likely captures more complex interdependencies and leads to improved realism. %An advantage of this approach is that dependencies between spawns and agents can be learned, which is a limitation of the nTPP-GMM.

\section{Conclusions}
We demonstrate that realistic crowd simulations can be orchestrated by spatio-temporal spawn dynamics, which include \textit{where} agents spawn, \textit{where} they will go, and \textit{when} they appear. More specifically, we propose nTPP-GMM to learn spatio-temporal spawn dynamics from real-world data, which allows to orchestrate crowd simulations where arbitrary policies can control agents. With nTPP-GMM, we propose a new learnable layer for realistic crowd simulations, which can be seamlessly integrated into existing crowd simulation frameworks. %Our approach moves beyond existing simulations based on only imitating microscopic agent behavior, traditional stochastic processes, or required expert knowledge for spawn dynamics. %By integrating neural Temporal Point Processes (nTPP) with Gaussian Mixture Models (GMM), our approach provides a controllable and data-driven approach for learning spatio-temporal agent dynamics in crowded scenes. 
Without the need for domain expert knowledge or any human interference, orchestrating a crowd with nTPP-GMM leads to the replication of real crowd scenarios, including diverse crowd flows and crowd densities, which can benefit several applications in urban planning, social robot navigation, and public safety in the future. We believe that this work paves the way for future research in realistic crowd simulations, and we hope that we can inspire more research focusing on the orchestration of crowds based on their spatio-temporal spawn dynamics.

\small
\section*{ACKNOWLEDGMENT}
This work has been partially funded by the LOEWE initiative (Hesse, Germany) within the emergenCITY centre and by the Federal Ministry of Education and Research (BMBF) grant 01$|$S17050. 

\bibliographystyle{IEEEtran}
\bibliography{IEEEabrv,main}

\end{document}